\documentclass{article}
\usepackage[final]{nips_2017}

\usepackage[utf8]{inputenc} 
\usepackage[T1]{fontenc}    
\usepackage[hyphens]{url}   
\usepackage{hyperref}       
\hypersetup{ hidelinks, }
\usepackage{booktabs}       
\usepackage{amsfonts}       
\usepackage{nicefrac}       
\usepackage{microtype}      
\usepackage[pdftex]{graphicx}
\usepackage{xcolor}

\newcommand{\ra}[1]{\renewcommand{\arraystretch}{#1}}

\usepackage{amsmath}

\title{Analyzing Language Learned by an \\Active Question Answering Agent}

\author{
  Christian Buck\\
  \texttt{cbuck@google.com}
  \And
  Jannis Bulian\\
  \texttt{jbulian@google.com}
  \And
  Massimiliano Ciaramita\\
  \texttt{massi@google.com}
  \And
  Wojciech Gajewski\\
  \texttt{wgaj@google.com}
  \And
  Andrea Gesmundo\\
  \texttt{agesmundo@google.com}
  \And
  Neil Houlsby\\
  \texttt{neilhoulsby@google.com}
  \And
  Wei Wang\\
  \texttt{wangwe@google.com}\\
  \\
  Google
}

\begin{document}

\maketitle
\begin{abstract}
  We analyze the language learned by an agent trained with reinforcement
learning as a component of the ActiveQA system~\citep{Buck:2017}. In
ActiveQA, question answering is framed as a reinforcement learning task
in which an agent sits between
the user and a black box question-answering system.
The agent learns to reformulate the user's questions to elicit the optimal answers.
It probes the system with many versions of a question
that are generated via a sequence-to-sequence question reformulation model,
then aggregates the returned evidence to find the best answer.
This process is an instance of \emph{machine-machine} communication.
The question reformulation model must adapt its
language to increase the quality of the answers returned,
matching the language of the question answering system.
We find that the agent does not
learn transformations that align with semantic intuitions but
discovers through learning classical information
retrieval techniques such as tf-idf re-weighting and stemming.
 \end{abstract}
\section{Introduction}
\begin{figure}
\centering
\includegraphics[width=0.5\linewidth]{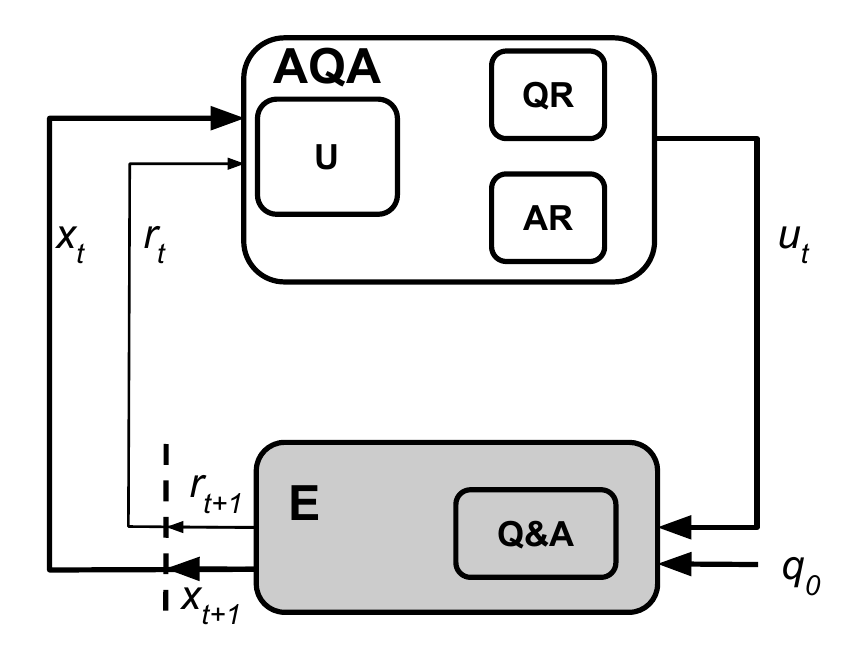}
\caption{An agent-environment framework for Active Question Answering.}
\label{fig:rl2}
\end{figure}

\citet{Buck:2017} propose a reinforcement learning framework for
question answering, called \emph{active question answering}
(ActiveQA), that aims
to improve answering by systematically perturbing
input questions (cf.~\citep{Nogueira:2017}).
Figure~\ref{fig:rl2} depicts the generic agent-environment framework.
The agent (AQA) interacts with the environment (E) in order to
answer a question ($q_0$). The environment includes a question
answering system (Q\&A), and emits observations and rewards.
A state $s_t$ at time $t$ is the sequence of observations and previous actions
generated starting from $q_0$: $s_t=x_0,u_0,x_1,\ldots,u_{t-1},x_t$,
where $x_i$ includes the question asked ($q_{i}$), the corresponding
answer returned by the QA system ($a_i$), and possibly additional
information such as features or auxiliary tasks. The agent
includes an action scoring component (U), which produced and action $u_t$ by
deciding whether to submit a new
question to the environment or to return a final answer. Formally,
$u_t\in \mathcal{Q}\cup \mathcal{A}$, where $\mathcal{Q}$ is the set
of all possible questions, and $\mathcal{A}$ is
the set of all possible answers. The agent relies on a question
reformulation system (QR), that provides candidate follow up
questions, and on an answer ranking system (AR), which scores the
answers contained in $s_t$.
Each answer returned is assigned a reward. The objective is to
maximize the expected reward over a set of questions.

\citet{Buck:2017} present a simplified version of this system
with three core components: a question reformulator, an off-the-shelf
black box QA system, and a candidate answer selection model.
The question reformulator is trained with policy gradient~\citep{Williams:1992}
to optimize the F1 score of the answers returned by the QA system to
the question reformulations in place of the original question.
The reformulator is implemented as a sequence-to-sequence
model of the kind used for machine translation~\citep{Sutskever:2014,bahdanau2014neural}.
When generating question reformulations, the action-space is equal to the size of
the vocabulary, typically $16k$ sentence
pieces.\footnote{\url{https://github.com/google/sentencepiece}}
Due to this large number of actions
we warm start the reformulation
policy with a monolingual sequence-to-sequence
model that performs generic paraphrasing.
This model is trained using the \emph{zero-shot}
translation technique~\citep{ZeroShot} on a large multilingual parallel
corpus~\citep{unv1}, followed by regular supervised learning on a
smaller monolingual corpus of questions~\citep{Fader:2013}.

The reformulation and selection models form a trainable agent that seeks the
best answers from the QA system. The reformulator proposes $N$
versions $q_i$ of the input question $q_0$ and passes them to the environment,
which provides $N$ corresponding answers, $a_i$. The selection model
scores each triple $(q_0,q_i,a_i)$ and returns the top-scoring
candidate.\footnote{For more details see~\citep{Buck:2017}.}

Crucially, the agent may only query
the environment with natural language questions.
Thus, ActiveQA involves a machine-machine communication process
inspired by the human-machine communication that takes place when
users interact with digital services during information seeking tasks.
For example, while searching for information on a search engine users
tend to adopt a keyword-like \emph{`queryese'} style of questioning.
The AQA agent proves effective at reformulating
questions on SearchQA~\citep{Dunn:2017}, a large dataset of complex
questions from the  \emph{Jeopardy!} game.
For this task BiDAF is chosen for the environment~\citep{Seo:2017},
a deep network built for QA which has produced state-of-the-art results.
Compared to a QA system that forms the environment
using only the original questions, AQA outperforms this baseline by a wide
margin, 11.4\% absolute F1, thereby reducing the gap between machine (BiDAF)
and human performance by 66\%.

Here we perform a qualitative analysis of this
communication process to better understand
what kind of language the agent has learned.
We find that while optimizing its reformulations
to adapt to the language of the QA system, AQA diverges
from well structured language in favour of less fluent,
but more effective, classic information retrieval (IR) query operations.
These include term re-weighting (tf-idf), expansion and
morphological simplification/stemming.
We hypothesize that the explanation of this behaviour is that
current machine comprehension tasks primarily require ranking of
short textual snippets, thus incentivizing relevance
more than deep language understanding.
 \section{Analysis of the Agent's Language}
\begin{figure}
\begin{center}
\includegraphics[trim={1cm 1cm 0 1cm},clip,width=1\linewidth]{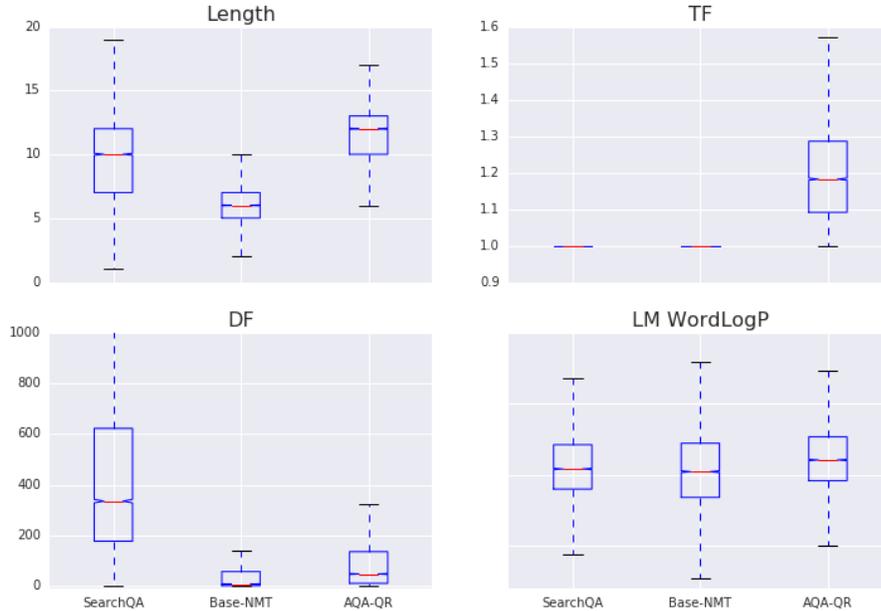}
\caption{Boxplot summaries of the statistics collected for all types of
  questions. Two-sample \emph{t}-tests performed on all pairs in
  each box confirm that the differences in means are significant $p<10^{-3}$.}
\label{fig:boxplot}
\end{center}
\end{figure}
We analyze input questions and reformulations on the $12k$ example
development partition of the SearchQA dataset.
Our goal is to gain insights on
how the agent's language evolves during training via policy gradient.
It is important to note that in the SearchQA dataset the original
\emph{Jeopardy!} clues have been preprocessed by lower-casing
and stop word removal. The resulting preprocessed clues that
form the sources (inputs) for the sequence-to-sequence reformulation model
resemble more keyword-based search queries than grammatical
questions. For example, the clue \emph{Gandhi was deeply influenced by
this count who wrote "War and Peace"} is simplified to \emph{gandhi
deeply influenced count wrote war peace}.

\subsection{The Language of SearchQA Questions}

Figure~\ref{fig:boxplot} summarizes statistics of the
questions and rewrites which may shed some light on how the language
changes.
The (preprocessed) SearchQA questions contain 9.6 words on average.
They contain few repeated terms, computed as the mean term frequency (TF) per
question.
The average is 1.03, but for most of the queries TF is
1.0, i.e.~no repetitions. We also compute the median
document frequency (DF) per query, where a document is the
context from which the answer is
selected.\footnote{We use the median instead of the mean to reduce the influence
of frequent outliers, such as commas, on the statistics. The mean DF is 460.}
DF gives a measure of how informative the question
terms are.

\subsection{The Language of the Base NMT Model}

We first consider the top hypothesis generated by the pre-trained
NMT reformulation system, before reinforcement learning (Base-NMT).
This system is trained with full supervision,
using a large multilingual and a small monolingual dataset.
The Base-NMT rewrites differ greatly from their sources.
They are shorter, 6.3 words
on average, and have even fewer repeated terms (1.01).
Interestingly, these reformulations are mostly syntactically well-formed
questions. For example, the
clue above becomes \emph{Who influenced count wrote war?}.\footnote{More
examples can be found in Appendix~\ref{app1}.}
Base-NMT improves structural language quality by properly reinserting
dropped function words and wh-phrases.
We also verified the increased
fluency by using a large language model and found that the Base-NMT
rewrites are 50\% more likely than the original questions.
The bottom right hand plot in Figure~\ref{fig:boxplot} summarizes the
language model distributions (LM~WordLogP).
The plot shows the average per-token language model negative log probabilities;
a lower score indicates greater fluency.
Although the distributions
overlap to a great extent due to the large variance across questions,
the differences in means are significant.

While more fluent, the Base-NMT rewrites involve rarer terms,
as indicated by the decrease in DF.
This is probably due to a domain mismatch
between SearchQA and the NMT training corpus.

\subsection{The Language of the AQA Agent}

We next consider the top hypothesis generated by the AQA question
reformulator (AQA-QR) after the policy gradient training.
The AQA-QR rewrites are those whose
corresponding answers are evaluated as
\emph{AQA Top Hyp.} in~\citet{Buck:2017}. Note, these single rewrites alone
outperform the original SearchQA queries by a small margin (+2\% on test).
We analyze the top hypothesis instead of the final output of the full
AQA agent to avoid confounding effects from the answer selection step.
These rewrites look
different from both the Base-NMT and the SearchQA
ones. For the example above AQA-QR's top hypothesis
is \emph{What is name gandhi gandhi influence wrote peace
peace?}. Surprisingly, 99.8\% start with the
prefix \emph{What is name}. The second most frequent is
\emph{What country is} (81 times), followed by \emph{What is is} (70) and
\emph{What state} (14). This is puzzling as it happens only for 9
Base-NMT rewrites, and never in the original SearchQA questions.
We speculate it might be related to the fact that
virtually all answers involve names,
of named entities (Micronesia) or generic concepts (pizza).
AQA-QR's rewrites are visibly less fluent than both the
SearchQA and the Base-MT counterparts.
In terms of language model probability they are less likely than both SearchQA
and Base-NMT.\footnote{To compute meaningful language model scores we
  remove the prefix ``What is name'' from all queries, because it
  artificially inflates the fluency measure, due to the high frequency
  unigrams and bigrams.}
However, they have more repeated terms (1.2 average TF),
are significantly longer (11.9) than the Base-NMT initialization
and contain more informative context terms (lower DF) than SearchQA questions.

Additionally, AQA-QR's reformulations contain
morphological variants in 12.5\% of cases.
The number of questions that
contain multiple tokens with the same stem doubles from SearchQA to
AQA-QR. Singular forms are preferred over plurals. Morphological
simplification is useful because it increases the chance
that a word variant in the question matches the context.

\section{Conclusions: Rediscovering IR?}

Recently, \citet{Lewis:2017} trained chatbots
that negotiate via language utterances in order to complete a
task. They report that the
agent's language diverges from human language if there is no
incentive for fluency in the reward function.
Our findings seem related.
The fact that the questions reformulated by AQA do not
resemble natural language is not due to the keyword-like
SearchQA input questions, because Base-NMT is capable of producing fluent
questions from the same input.

AQA learns
to re-weight terms by focusing on informative (lower DF) terms
while increasing term frequency (TF) via duplication. At the same time
it learns to modify surface forms in ways akin to stemming and
morphological analysis.
Some of the techniques seem to
adapt also to the specific properties of current deep QA architectures
such as character-based modelling and attention.
Sometimes AQA learns to generate
semantically nonsensical, novel, surface term variants; e.g., it
might transform the adjective \emph{dense} to \emph{densey}. The only
justification for this is that such forms can be still exploited by the
character-based BiDAF question encoder.
Finally, repetitions can directly increase
the chances of alignment in the attention components.

We hypothesize that there is no
incentive for the model to use human language
due to the nature of the task.
AQA learns to ask BiDAF questions by
optimizing a language that increases the likelihood of BiDAF
\emph{extracting} the right answer.
\citet{Jia:2017} argue that reading comprehension
systems are not capable of significant language understanding and
fail easily in adversarial settings.
We suspect that current machine comprehension tasks involve mostly
simple pattern matching and relevance modelling. As a consequence
deep QA systems behave as sophisticated
ranking systems trained to sort snippets of text from the context.
As such, they resemble document retrieval systems
which incentivizes the (re-)discovery
of IR techniques that have been successful for
decades~\citep{Baeza-Yates:1999}.
 
\newpage
\bibliographystyle{abbrvnat}
{\small\bibliography{nips}}
\newpage
\appendix
\section{Examples}
\begin{footnotesize}
\begin{table*}[h!]
  \centering
  \ra{1.0}
  \begin{tabular}{l|p{12cm}}
    \toprule
    Jeopardy! & People of this nation AKA Nippon wrote with a brush, so painting became the preferred form of artistic expression \\
    SearchQA & people nation aka nippon wrote brush , painting became preferred form artistic expression \\
    Base-NMT & Aka nippon written form artistic expression?\\
    AQA-QR & What is name did people nation aka nippon wrote brush expression?\\
    AQA-full & people nation aka nippon wrote brush , painting became preferred form artistic expression \\
    \midrule
    Jeopardy! & Michael Caine \& Steve Martin teamed up as Lawrence \& Freddy, a couple of these, the title of a 1988 film\\
    SearchQA & michael caine steve martin teamed lawrence freddy , couple , title 1988 film \\
    Base-NMT & Who was lawrence of michael caine steve martin? \\
    AQA-QR & What is name is name is name michael caine steve martin teamed lawrence freddy and title 1988 film?\\
    AQA-full & What is name is name where name is name michael caine steve martin teamed lawrence freddy and title 1988 film key 2000 ? \\
    \midrule
    Jeopardy! & Used underwater, ammonia gelatin is a waterproof type of this explosive \\
    SearchQA  & used underwater , ammonia gelatin waterproof type explosive\\
    Base-NMT & Where is ammonia gelatin waterproof?\\
    AQA-QR & What is name is used under water with ammonia gelatin water waterproof type explosive? \\
    AQA-full & used underwater , ammonia gelatin waterproof type explosive \\
    \midrule
    Jeopardy! & The Cleveland Peninsula is about 40 miles northwest of Ketchikan in this state\\
    SearchQA & cleveland peninsula 40 miles northwest ketchikan state \\
    Base-NMT & The cleveland peninsula 40 miles? \\
    AQA-QR & What is name is cleveland peninsula state northwest state state state?\\
    AQA-full & What is name are cleveland peninsula state northwest state state state ?\\
    \midrule
    Jeopardy! & Tess Ocean, Tinker Bell, Charlotte the Spider \\
    SearchQA & tess ocean , tinker bell , charlotte spider \\
    Base-NMT & What ocean tess tinker bell? \\
    AQA-QR & What is name tess ocean tinker bell link charlotte spider?\\
    AQA-full & What is name is name tess ocean tinker bell spider contain charlotte spider contain hump around the world winter au to finish au de mon moist    \\
    \midrule
    Jeopardy! & During the Tertiary Period, India plowed into Eurasia \& this highest mountain range was formed\\
    SearchQA & tertiary period , india plowed eurasia highest mountain range formed \\
    Bas-NMT & What is eurasia highest mountain range? \\
    AQA-QR & What is name were tertiary period in india plowed eurasia? \\
    AQA-full & tertiary period , india plowed eurasia highest mountain range formed\\
    \midrule
    Jeopardy! & The melody heard here is from the opera about Serse, better known to us as this "X"-rated Persian king\\
    SearchQA & melody heard opera serse , better known us x rated persian king\\
    Base-NMT& Melody heard opera serse thing?\\
    AQA-QR & What is name melody heard opera serse is better persian king?\\
    AQA-full & What is name is name melody heard opera serse is better persian king persian K ?\\
    \midrule
    Jeopardy! & A type of humorous poem bears the name of this Irish port city \\
    SearchQA & type humorous poem bears name irish port city \\
    Base-NMT & Name of humorous poem bears name? \\
    AQA-QR & What is name is name humorous poem poem bear city city city? \\
    AQA-full & What is name is name were humorous poem poem bears name city city city ? \\
   \bottomrule
  \end{tabular}
  \label{app1}
\end{table*}

\end{footnotesize}
 \end{document}